\documentclass[10pt,twocolumn,letterpaper]{article}
\usepackage{url}
\usepackage{cvpr}
\usepackage{times}
\usepackage{epsfig}
\usepackage{graphicx}
\usepackage{amsmath}
\usepackage{amssymb}

\usepackage{booktabs}
\usepackage{multirow}
\usepackage{subfig}

\usepackage{algorithm}
\usepackage{algorithmic}
\usepackage{textcomp}
\usepackage{gensymb}
\usepackage{amsmath}
\usepackage{amssymb}
\usepackage{booktabs}
\usepackage{threeparttable}
\usepackage[table,xcdraw]{xcolor}
\usepackage{comment}
\usepackage{arydshln}
\usepackage{makecell}
\usepackage[super]{nth}
\usepackage{threeparttable}
\usepackage[super]{nth}

\let\svthefootnote\thefootnote
\newcommand\blankfootnote[1]{%
  \let\thefootnote\relax\footnotetext{#1}%
  \let\thefootnote\svthefootnote%
}

\newcommand\blfootnote[1]{%
  \begingroup
  \renewcommand\thefootnote{}\footnote{#1}%
  \addtocounter{footnote}{-1}%
  \endgroup
}

\usepackage{cuted}
\usepackage{capt-of} 
\usepackage[table]{xcolor} 

\usepackage[breaklinks=true,bookmarks=false,urlcolor=magenta,linkcolor=blue,colorlinks]{hyperref}

\definecolor{Gray}{gray}{0.95}
\newcolumntype{g}{>{\columncolor{Gray}}c}

\cvprfinalcopy 

\def\cvprPaperID{****} 

\begin{document}

\title{Multiverse Transformer: 1st Place Solution for
Waymo Open Sim Agents Challenge 2023}

\author{Yu Wang \quad Tiebiao Zhao \quad Fan Yi  \\
Pegasus\\
{\tt\small yuwangrpi@gmail.com}
}

\twocolumn[{%
\renewcommand\twocolumn[1][]{#1}%
\maketitle
\begin{center}
    \centering
    \captionsetup{type=figure}
    \includegraphics[width=\textwidth]{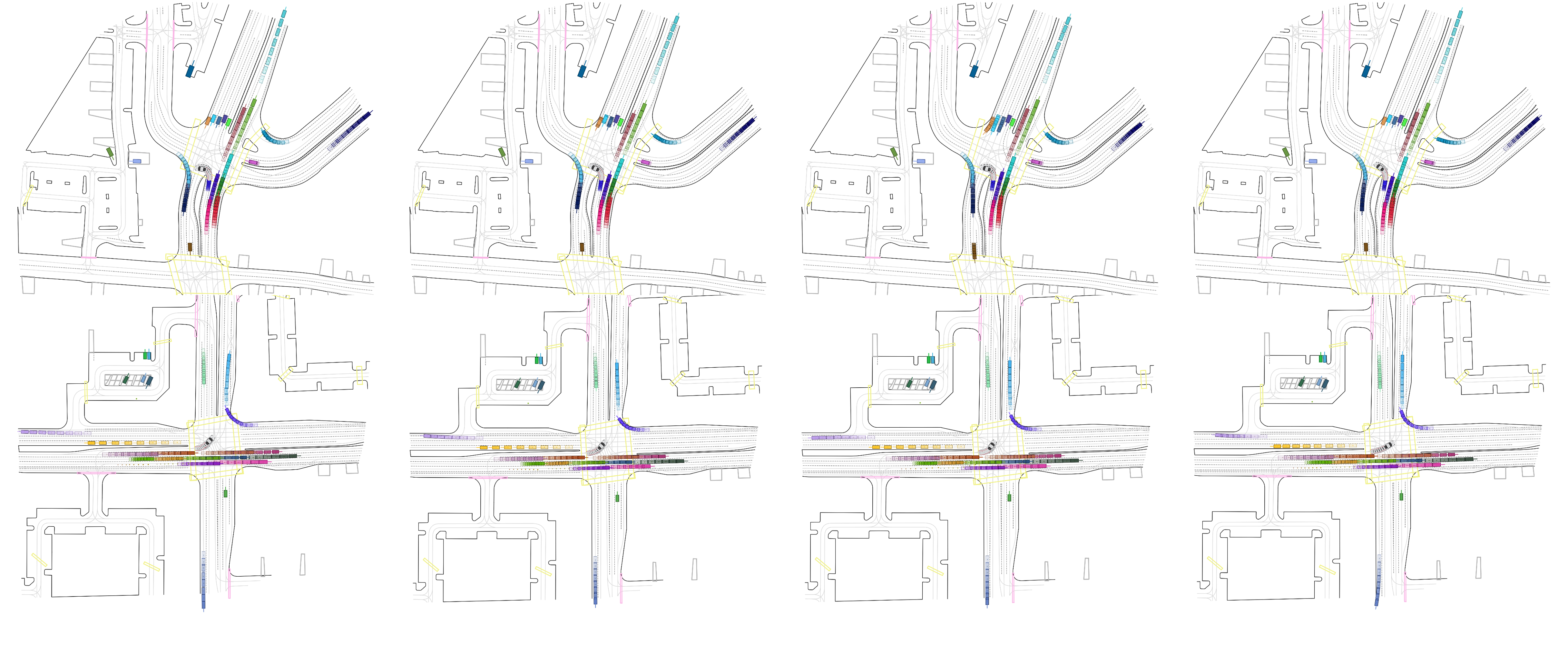}
    \captionof{figure}{Simulations of agents produced by MVTA. 4 parallel universes are visualized for each of the two scenes, in which the autonomous driving vehicle (ADV) uniquely represented by a car icon for easy identification is undertaking a U-turn maneuver (top) or making a left turn (bottom). For illustrative  purposes, eight timesteps are shown for each agent, and time progression is encoded in the opacity of the boxes.}
    \label{fig:examples}
\end{center}
}]

\begin{abstract}
This technical report presents our 1\textsuperscript{st} place solution for the Waymo Open Sim Agents Challenge (WOSAC) 2023. Our proposed \textbf{M}ulti\textbf{V}erse \textbf{T}ransformer for \textbf{A}gent simulation (\textbf{MVTA}) effectively leverages transformer-based motion prediction approaches, and is tailored for closed-loop simulation of agents. In order to produce simulations with a high degree of realism, we design novel training and sampling methods, and implement a receding horizon prediction mechanism. In addition, we introduce a variable-length history aggregation method to mitigate the compounding error that can arise during closed-loop autoregressive execution. On the WOSAC, our \textbf{MVTA} and its enhanced version \textbf{MVTE} reach a realism meta-metric of 0.5091 and 0.5168, respectively, outperforming all the other methods on the leaderboard. \blfootnote{ Project page: \url{https://multiverse-transformer.github.io/sim-agents/}}

\end{abstract}

\begin{figure*}
\centering
\includegraphics[width=0.93\textwidth]{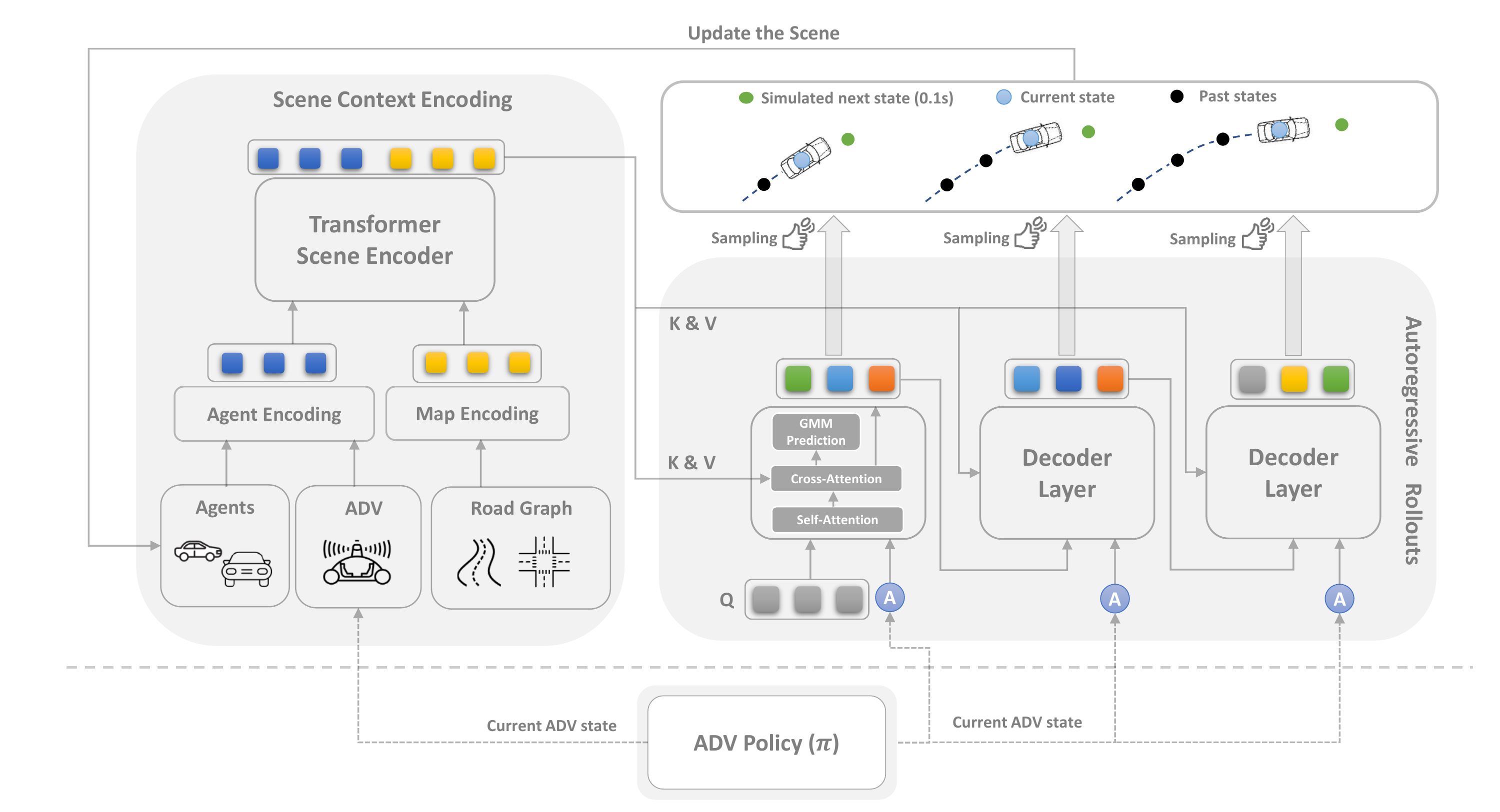}
\caption{\textbf{Main Architecture of the MVTA}. The world agents, ADV and road graph inputs are processed by transformer-based encoding steps to generate enhanced scene context features. In the decoding step, rollout is executed in an autoregressive manner. The ADV operates at 0.1s intervals, and concurrently, each world agent decoder also simulates the forthcoming states at the same 0.1s interval. \textit{Q} denotes the query content feature, while \textit{K} and \textit{V} stand for the keys and values, respectively. The coin flip icon indicates sampling at each timestep.}
\label{fig:architecture}
\end{figure*}

\section{Introduction}
The simulation of traffic agents is an integral element for evaluating self-driving systems, facilitating rapid development and ensuring safety~\cite{suo2021trafficsim}.  WOSAC~\cite{montali2023waymo} is the first public benchmark for the evaluation of simulation agents in the domain of autonomous driving, introducing new evaluation metrics and leveraging large-scale real-world logged data~\cite{ettinger2021large} with a diverse set of scenarios and agent behaviors.

Recent advancements in traffic agent simulators~\cite{suo2021trafficsim,bergamini2021simnet, feng2023trafficgen, zhong2022guided, igl2022symphony, rempe2022generating, xu2022bits} have shown a notable shift towards learning from logged real-world driving data and data-driven generative models conditioned on the scene context, rather than relying on traditional heuristic-based models encoding traffic rules. Our proposed simulator also falls within the learning-based generative model paradigm. Specifically, we leverage state-of-the-art motion prediction models~\cite{shi2023motion, nayakanti2022wayformer} and adapt them to an autoregressive closed-loop agent simulator. 

However, for autoregressive models, the error can accumulate over time, as future state of agents is heavily dependent on the immediate past. This can potentially lead to drift, where the predictions become increasingly inaccurate over time. 

Inspired by TrafficSim~\cite{suo2021trafficsim} and MTR~\cite{shi2023motion}, we propose a novel closed-loop simulation framework based on transformer encoder and decoder, and further enhance the realism of the simulations through the development of novel training and sampling strategies, as well as the receding horizon prediction and variable-length history aggregation methods. Several example simulations generated by MVTA are depicted in Figure~\ref{fig:examples}.

\section{Related Work}
\noindent \textbf{Multi-modal motion prediction}.
In the motion prediction literature, there are agent-centric prediction algorithms~\cite{nayakanti2022wayformer, shi2023motion}, as well as scene-centric and joint multi-agent~\cite{girgis2022latent, ngiam2022scene, yuan2021agentformer, luo2022jfp} prediction methods. Most recent prediction methods adopt encoder-decoder Transformer networks~\cite{nayakanti2022wayformer, shi2023motion, achaji2022pretr}. There is also a trend of employing diffusion in the literature for predicting trajectories through the denoising process~\cite{rempe2023trace, jiang2023motiondiffuser}. 

Most motion prediction methods are open-loop, in the sense that the whole trajectory is produced in one-shot independently. However, in this challenge, traffic agent simulation needs to be simulated in a closed-loop autoregressive manner at 0.1s intervals. Additionally, the method must differentiate the world agents and the ADV and factorize their joint distribution for conditional independence.

\noindent \textbf{Traffic agent simulation}.
TrafficSim~\cite{suo2021trafficsim} utilizes real-world logged data to mimic a broad range of human driving behaviors, and leverages an implicit latent variable model~\cite{casas2020implicit} to generate socially-consistent plans for all traffic actors jointly. TrafficGen~\cite{feng2023trafficgen} places agents in the scene based on the learned distribution and simulates their future states. CTG~\cite{zhong2022guided} developed a conditional diffusion model that allows user to control over trajectory properties while maintaining realism.

The WOSAC requires the simulator to be agnostic to the choice of ADV policy, therefore it can be swapped with arbitrary ADV policy or planner. Both ADV and environment agent models need to obtain multiple modes in order to perform well~\cite{montali2023waymo}.

\section{MultiVerse Transformer Agent Simulator (MVTA)}
\noindent \textbf{Problem formulation}. Given the scene context, including map and past positions of the agents (\ie, world agents and ADV), the goal is to simulate new states of the agents at 0.1s intervals for the upcoming $T=80$ timesteps (\ie, a 8s episode). There are two constraints: 1) the simulator must be closed-loop and run in autoregressive manner; 2) the joint distribution involving the world agents and ADV must be factorized into two conditionally independent components, to ensure that the ADV component can be replaced with any arbitrary policy or planner. 

\begin{figure}
  \centering
   \includegraphics[width=0.35\textwidth]{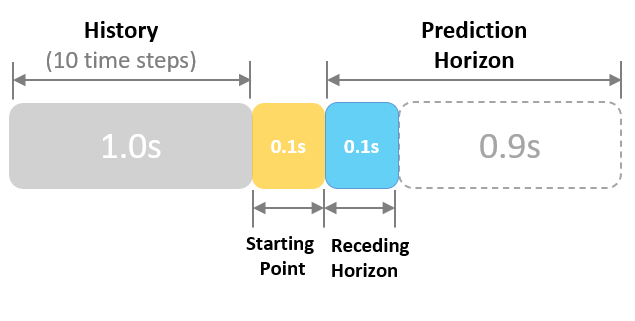}
  \caption{Illustrating the receding horizon. Even though predictions are made for the next 1s, only the waypoint of the initial 0.1s is utilized, with the remaining prediction being discarded.}
  \label{fig:receding_horizon}
\end{figure}

\subsection{Network Architecture}
\noindent \textbf{Main architecture}. The main architecture of MVTA is illustrated in Figure~\ref{fig:architecture}. Scene context features are obtained by processing the world agents, ADV, and map data through polyline and transformer-based encoding steps. The transformer decoder layer takes the scene context features and queries as input and unrolls the agent states for the next timestep. This architecture is implemented to fulfill the requirement of executing closed-loop simulations at 0.1s intervals. Current state of the ADV is also used as input to the decoder layer so the environment agents can react to it. Query content feature output by each decoder layer is used as the input for the subsequent decoder layer. The motion prediction head, based on Gaussian Mixture Model (GMM), outputs multi-modal trajectory predictions. To sample the state from the multi-modal prediction, we either pick the maximum-likelihood trajectory or randomly sample from the top-\textit{k} trajectories with the highest likelihood. In our implementation, we leverage the same architecture for the ADV policy, but it can be swapped with any policy or planner.

\noindent \textbf{Transformer-based scene encoder}.
Given the agent-centric scene inputs (\ie, agents, ADV, and road graph), we utilize their vector representation \cite{gao2020vectornet}, and adopt polyline encoders consisting of a multi-layer perceptron network (MLP) followed by maxpooling~\cite{shi2023motion}. As in~\cite{shi2023motion}, the agent input is represented as the agent history motion state (\ie, agent position, size, heading and velocity) with a one-hot category mask, while the map input consists of the position and direction of each polyline point and the polyline type. The polyline encoders produce agent features $A_{past}\in\mathbb{R}^{N_{a} \times D}$ and map features $M_{past}\in\mathbb{R}^{N_{m} \times D}$, where $N_{a}$ and $N_{m}$ are the number of agents and map polylines, respectively, and $D$ is the feature dimension. A scene context transformer encoder leverages local self-attention~\cite{shi2023motion} to produce enhanced scene context features that serve as inputs of the subsequent decoder network.

\noindent \textbf{Autoregressive transformer decoder}.
The autoregressive decoding consists of a group of transformer decoder layers. Each decoder layer has a self-attention component, and a cross-attention component that attends to the scene context features, and a GMM prediction module that produces multi-modal predictions. Each Gaussian component is represented as $(\mu_{x}, \mu_{y}, \sigma_{x}, \sigma_{y}, \rho)$ and predicted with a probability $p$. The motion prediction head also predicts the velocity $(v_{x}, v_{y})$ and heading angle $(sin(\theta), cos(\theta))$ of each agent for the next timestep. We adopt the motion query pair design in~\cite{shi2023motion}. There are a total of 64 queries, corresponding to the 64 motion modes, each associated with an intention point. 

\noindent \textbf{Receding horizon}. The next 0.1s state is simulated by sampling the multi-modal predictions output by the decoder layer. However, each decoder layer is trained to predict a 1s trajectory, and we adopt a receding horizon solution in which only the initial 0.1s is utilized, as illustrated in Figure~\ref{fig:receding_horizon}. The benefits of longer prediction horizon include the promotion of multi-modal diversity, reduction of compounding error and also more flexibility in inference setup.

\begin{figure}
  \centering
   \includegraphics[width=0.35\textwidth]{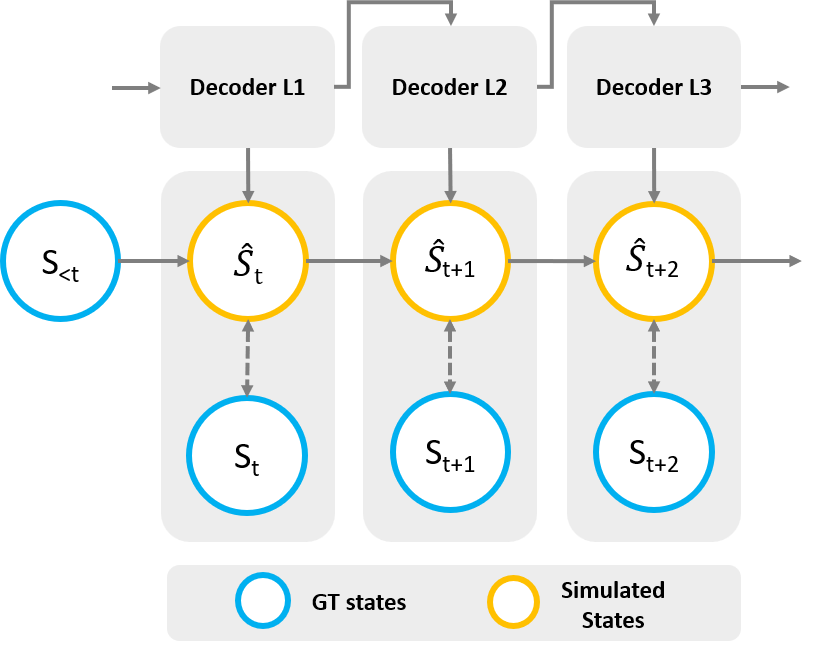}
  \caption{Training loss is calculated at each timestep. Each decoder layer produces multi-modal predictions for one timestep.}
  \label{fig:training_loss}
\end{figure}

\noindent \textbf{Scene context update}. The states output by the decoder layer at each timestep are used to update the scene context features. However, the design of our decoder allows running the scene encoding at specified periodic intervals, rather than at every simulation step. Alternatively, to keep the scene context features updated when predicting the next timestep, we implement two modifications to the network. Firstly, current position of the ADV is used as input to the decoder layer. Similar to the static intention query in~\cite{shi2023motion}, sinusoidal position encoding and MLP are applied, and the resulting position embedding is added to the query content. Secondly, we update the agent features with additional features encoding the current positions of the other agents $A = MLP([A_{past}, A_{current}])$.

\subsection{Training}
Our simulation model is trained end-to-end in a closed-loop manner similar to that used in~\cite{suo2021trafficsim}. As shown in Figure~\ref{fig:training_loss}, supervision is provided at each decoder layer, and losses are computed for each timestep.

\noindent \textbf{Training samples}. The training samples are generated to accommodate variable lengths of past history, as opposed to adhering to a fixed length of 1.1s. Specifically, for each 9.1s training trajectory, we randomly pick a point to separate the trajectory to history and future components. This way, more training samples can be generated from each ground-truth trajectory. Moreover, this facilities the trajectory history aggregating mechanism in our inference step.

\noindent \textbf{Training losses}. We use \textit{L}1 loss for regressing the agent velocity and heading angles, and the Gaussian regression loss implemented based on the negative log-likelihood loss to maximize the likelihood of ground-truth trajectory~\cite{shi2023motion}. 

At each timestep the loss can be formulated as:
\begin{equation}
\mathcal{L}_{NLL} = -{\rm log}\mathcal{N}(S_{x} - \mu_{x}, \sigma_{x}; S_{y} - \mu_{y}, \sigma_{y}; \rho)
\end{equation}
where $S_{x}, S_{y}$ is the waypoint of the ground-truth trajectory at this timestep, and $(\mu_{x}, \mu_{y}, \sigma_{x}, \sigma_{y}, \rho)$ represent the parameters of the selected positive Gaussian component.

We calculate the final loss based on the weighted sum of the $\mathcal{L}_{NLL}$ loss and the \textit{L}1 losses of velocity and heading angles.
\begin{equation}
\mathcal{L}_{total} = \sum_{t}\lambda_{1}\mathcal{L}_{NLL}^{t} + \lambda_{2}\mathcal{L}_{Vel}^{t} + \lambda_{3}\mathcal{L}_{\theta}^{t}
\label{eq:final_loss}
\end{equation}

During the challenge, we also made an attempt to implement a simplified version of the collision avoidance loss~\cite{suo2021trafficsim}, however the preliminary result did not indicate any improvement in terms of realism metric, and we leave it to future studies.

\subsection{Inference}
\begin{figure}
  \centering
  \subfloat[]{
    \includegraphics[width=0.22\textwidth]{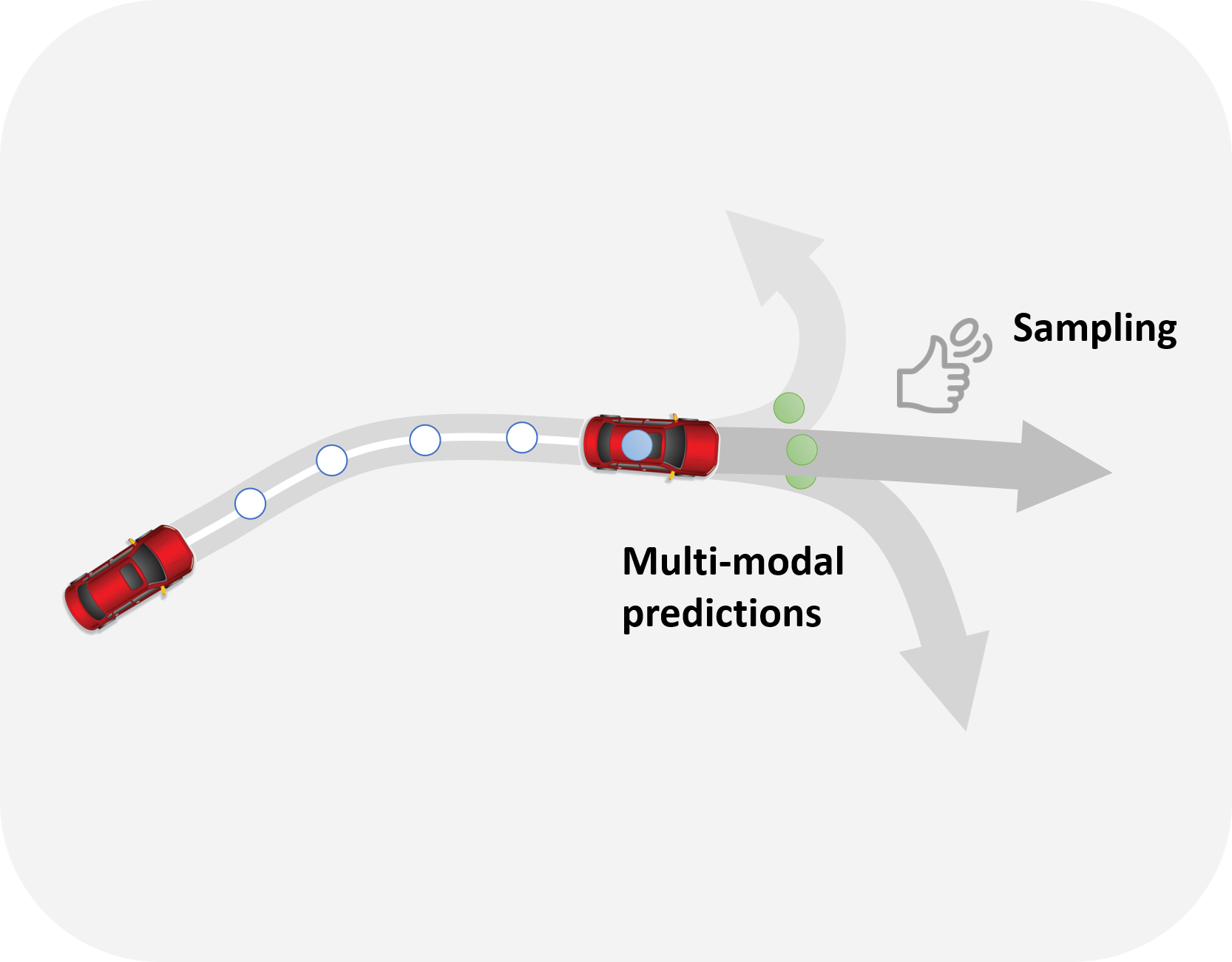}
    \label{fig:top_k_sampling}
  }
  \hfill
  \subfloat[]{
    \includegraphics[width=0.22\textwidth]{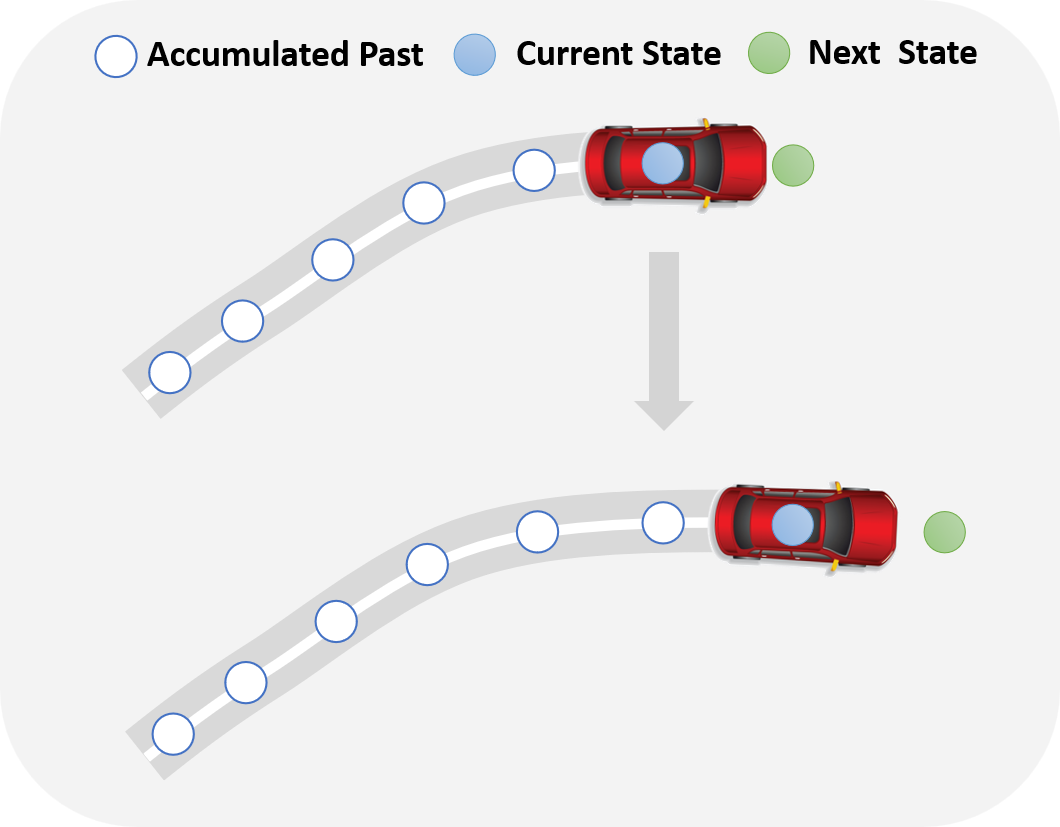}
    \label{fig:aggregating_past_trajectory}
  }
  \caption{(a) Top-\textit{k} sampling. (b) Aggregating new waypoint to the past trajectory.}
\end{figure}

\noindent \textbf{Top-\textit{k} sampling}. During model inference, each simulation step produces multi-modal trajectories. There are two alternatives available for trajectory sampling. The first approach is to select the maximum-likelihood trajectory, while the second is to randomly choose among the top-\textit{k} (\eg, \textit{k}=3) trajectories (Figure~\ref{fig:top_k_sampling}) with the highest likelihood.  The first one, while producing accurate trajectory, tends to yield less varied trajectories. On the other hand, opting for the top-\textit{k} trajectories encourages diversity but is susceptible to the compounding error and could generate trajectories with unrealistic kinematic motions or even drift, As a result, we employ the top-\textit{k} sampling at periodic intervals during the simulation steps to strike a balance between realism and diversity. 

\begin{figure*}
  \centering
   \includegraphics[width=0.88\textwidth]{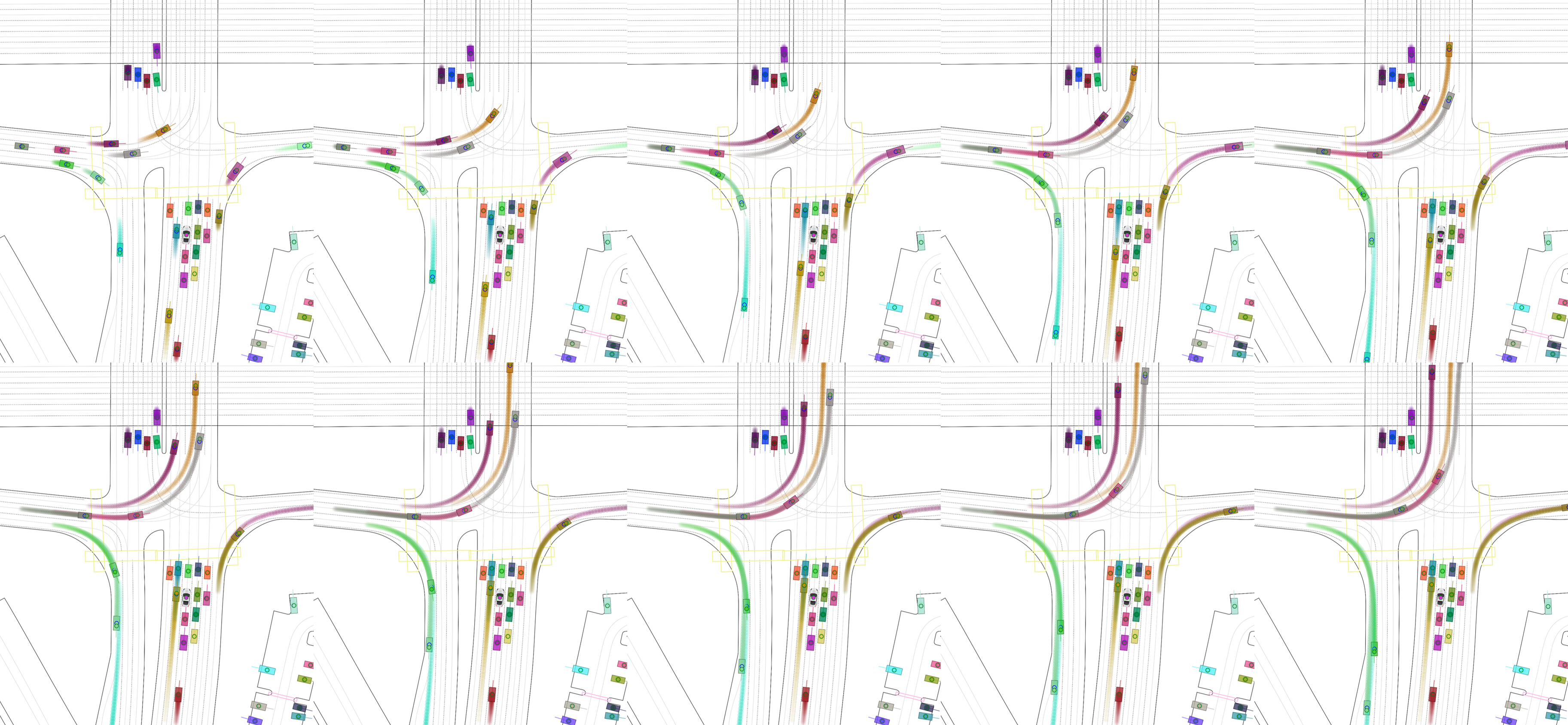}
  \caption{An example of the autoregressive rollout. The context history, current state (blue-edged circle) and next simulated waypoint (green-edged circle) are visualized for each agent.}
  \label{fig:autoregressive_rollouts}
\end{figure*}

\noindent \textbf{Variable-length history aggregation}. Instead of using fixed 1.1s history, we continuously aggregate the past history as the agent state unrolls overtime (Figure~\ref{fig:aggregating_past_trajectory}), and use the aggregated history trajectory for the scene context encoding in the next simulation step. The motivation is two-fold, firstly, our training process also uses variable-length history. Secondly, we aim to enhance the stability of the trajectory simulation, thereby reducing the potential for compounding error. One example showcasing the autoregressive rollout is shown in Figure~\ref{fig:autoregressive_rollouts}.

\section{Experimental Evaluation}
\begin{table*}[t]
\centering
\resizebox{\linewidth}{!}{
\begin{threeparttable}
\begin{tabular}{c>{\columncolor[gray]{0.95}}ccccccccccc}
\Xhline{4\arrayrulewidth}
  \textbf{WAYMO} & META METRIC & \multicolumn{4}{c}{KINEMATIC} & \multicolumn{3}{c}{INTERACTIVE} & \multicolumn{2}{c}{MAP} \\  \cmidrule(lr){2-2} \cmidrule(lr){3-6} \cmidrule(lr){7-9} \cmidrule(lr){10-11}
  \textbf{LEADERBOARD} &  \textbf{REALISM} & LINEAR & LINEAR & ANG. & ANG. & DIST. & COLLISION & TTC & DIST. & OFFROAD & minADE \\ 
 &  & SPEED & ACCEL. & SPEED & ACCEL. & TO OBJ. & & & TO ROAD & & ($\downarrow$) \\ \hline

\Xhline{4\arrayrulewidth}

\textbf{MVTE} (ours) & \textbf{0.5168} & 0.4426 & 0.2218 & \textbf{0.5353} & \textbf{0.481} & \textbf{0.382} & 0.4509 &\textbf{ 0.832} &\textbf{ 0.6641} & \textbf{0.6409} & 1.677\\
\textbf{MVTA} (ours) & 0.5091 & 0.4365 & 0.22 & 0.533 & 0.4805 & 0.3729 & 0.4359 & 0.8298 & 0.6545 & 0.6288 & 1.8698\\ \Xhline{2\arrayrulewidth}
MTR+++ & 0.4697 & 0.4119 & 0.1066 & 0.4838 & 0.4365 & 0.3457 & 0.4144 & 0.7969 & 0.6545 & 0.577 & 1.6817\\
CAD & 0.4321 & 0.3464 & \textbf{0.2526} & 0.4327 & 0.311 & 0.33 & 0.3114 & 0.7893 & 0.6376 & 0.5397 & 2.3146\\
multipath & 0.424 & 0.4318 & 0.2304 & 0.0193 & 0.0355 & 0.3493 & \textbf{0.4854} & 0.8111 & 0.6372 & 0.613 & 2.0517\\
sim\_agents\_tutorial & 0.3941 & 0.3143 & 0.1738 & 0.4785 & 0.4631 & 0.2641 & 0.2671 & 0.7709 & 0.5575 & 0.4111 & 3.6198\\
QCNeXt & 0.392 & \textbf{0.4773} & 0.2424 & 0.3252 & 0.1987 & 0.3759 & 0.3244 & 0.7569 & 0.6099 & 0.36 & \textbf{1.083}\\
sim\_agents\_tutorial & 0.3201 & 0.3826 & 0.0999 & 0.0318 & 0.0391 & 0.2909 & 0.336 & 0.7549 & 0.5251 & 0.3804 & 3.108\\
linear\_extrapolation\_baseline\_tutorial & 0.2576 & 0.0745 & 0.1659 & 0.0187 & 0.0348 & 0.2221 & 0.2211 & 0.7551 & 0.479 & 0.3352 & 7.5148\\

\Xhline{4\arrayrulewidth}
\end{tabular}
\end{threeparttable}
}
\caption{\textbf{WOSAC Leaderboard}. Realism meta-metric is the primary metric for ranking the methods. Our simulator reached the highest meta-metric of 0.5168 among all the methods on the leaderboard.}
\label{table:leaderboard}
\end{table*}

\begin{figure*}
    \centering
    \subfloat{\includegraphics[width=0.70\textwidth]{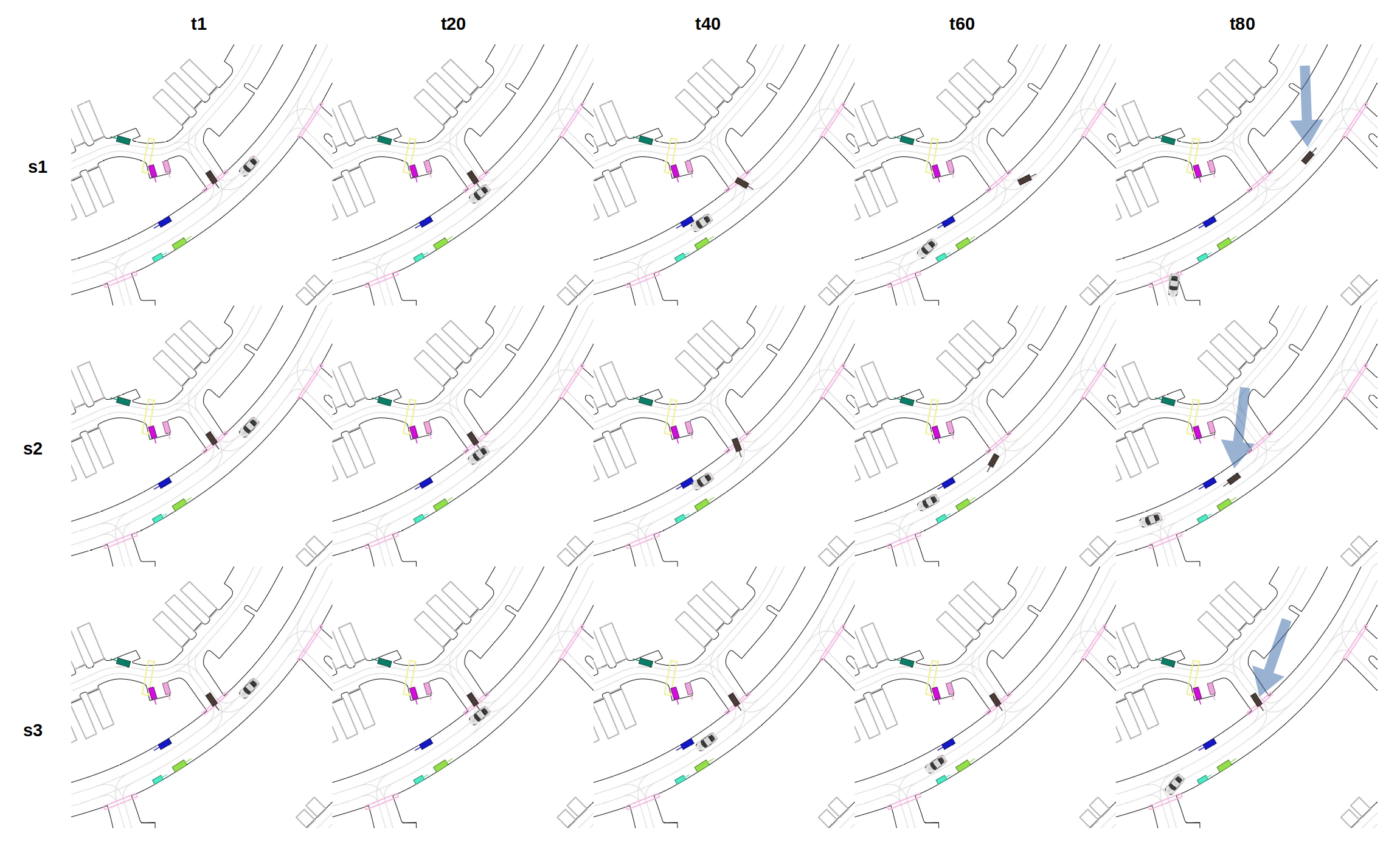}}\\
    \caption{Three simulations of a scene, in which a vehicle waiting to get onto the main road. The vehicle turns left or right, or keeps waiting for the right time to go. Additionally, the ADV also demonstrates multi-modal behavior, either proceeding straight or making a left turn. Five timesteps are rendered for each simulation.}
    \label{fig:qualitative_example1}
\end{figure*}

\subsection{Dataset and Metrics}
We use the Waymo Open Motion Dataset (WOMD)~\cite{ettinger2021large} v1.2.0 release in our experiments. There are a total of 486,995, 44,097, and 44,920 scenarios in the training, validation, and test set, respectively. Each scenario in the training and validation sets comprises of 11 observations for history and 80 observations from 8 seconds of future data, therefore the total duration of each scenario is 9.1 seconds. 

The task is to simulate up to 128 agents including the ADV, and generate 80 simulation steps (8s) for each agent in a 0.1s sampling interval, and in an autoregressive and reactive manner~\cite{montali2023waymo}. 32 simulations are required for each agent to be simulated.

There are three object types (vehicles, cyclists, and pedestrians), and their $x/y/z$ centroid coordinates and heading need to be simulated. There is no need to simulate the size of each agent since it stays constant. In our experiments, we keep the $z$ value the same as the starting state. The challenge does not enforce any motion model, and therefore there are no kinematic constraints. 

The main evaluation metric is the realism meta-metric, aggregating a group of component metrics including kinematic, interactive and map-based metrics. For more details, please refer to~\cite{montali2023waymo}.

\subsection{Implementation and Simulation Setup}
Table~\ref{table:hyperparameters} summarizes the hyperparameters of different modules used in our implementation.

\noindent \textbf{Training details}. The simulation model is trained end-to-end for all three agent types, using AdamW optimizer for 30 epochs. The learning rate is set to 0.0001.  We set the loss weights $\lambda_{1}, \lambda_{2}, \lambda_{3}$ in Equation~\eqref{eq:final_loss} to 1.0, 0.5, 0.5, respectively. Similar to~\cite{shi2023motion}, we use 64 motion query pairs based on 64 intention points learned by running $k$-means clustering algorithm on the future 1s waypoints of the training trajectories. A set of 64 intention points is obtained for each object category.

\noindent \textbf{Batch inference and optimization}. There are a total of 44,920 scenes in the test set, and each scene requires running the model inference for $32 \times T$ times to generate the 32 simulations for a group of agents. As such, we implemented batch inference to speed up the simulation process. Given that our model design supports periodic updates of the scene context features, the inference speed can be further optimized by running the scene context encoder every few timesteps (\eg, 0.5s) and running several decoder layers (\eg, the first 5).

\begin{table}[ht]
    \centering
    \rowcolors{2}{gray!10}{white}
    \resizebox{0.32\textwidth}{!}{%
    \begin{tabular}{llc}
    \Xhline{4\arrayrulewidth}
        \rowcolor{gray!20}
        \textbf{Module} & \textbf{Hyperparameters} & \textbf{Values} \\ \hline\hline
        \textbf{Scene MLP} & No. Channels-Agent & 256 \\
                        & No. Layers-Agent & 3 \\
                        & No. Channels-Map & 64 \\
                        & No. Layers-Map & 5 \\
        \textbf{Encoder} & Hidden Feature Dim. & 256/384 \\
                & No. Encoder Layers & 6 \\
                & No. Attention Head & 8 \\
        \textbf{Decoder} & Hidden Feature Dim.-Agent & 512/768 \\
                & Hidden Feature Dim.-Map & 256/384 \\
                & No. Decoder Layers & 10 \\
                & No. Attention Head & 8 \\
                & No. Motion Modes & 64 \\
        \textbf{Training} & Learning rate & 0.0001 \\
                & No. Epochs & 30 \\
                & Loss weights & 1.0, 0.5, 0.5 \\
        \Xhline{4\arrayrulewidth}
    \end{tabular}%
    }
    \caption{Hyperparameters of different modules in MVTA.}
    \label{table:hyperparameters}
    \vspace{-2pt}
\end{table}

\noindent \textbf{MVTE}. We explore the design space of the MVTA model and  trained 3 variants of the model by increasing the number of hidden feature dimension in the encoder (\eg, 384 as opposed to 256) and decoder (\eg, 768 as opposed to 512). In the enhanced MVTE solution, model is randomly sampled to generate each simulation, encouraging more diversity in the resulting simulations.

\subsection{Experimental Results}
\noindent \textbf{WOSAC 2023 leaderboard}. On the WOSAC leaderboard\footnote{\url{https://waymo.com/open/challenges/2023/sim-agents/}}, the realism meta-metric is the official primary metric used for ranking the methods. The minADE metric is also calculated but it is primarily used for evaluating motion prediction methods. The official baseline extrapolates the trajectory of an agent using the last heading and speed logged in the provided history ~\cite{montali2023waymo}. For more baselines based on Wayformer ~\cite{nayakanti2022wayformer} on the validation set, please refer to ~\cite{montali2023waymo}.

The leaderboard is shown in Table~\ref{table:leaderboard}. On the test set, our MVTA reaches a realism meta-metric of 0.5091  and our MVTE further improves the meta-metric to 0.5168, ranking the 1\textsuperscript{st} place in the challenge. Notably, it also has the highest scores in component metrics except the linear and collision metrics.  

Note in Table~\ref{table:leaderboard} that minADE does not always correlate with the ranking, as the method achieving the lowest minADE has lower realism meta metric compared to other methods.

\begin{figure*}
    \centering
    \subfloat{\includegraphics[width=0.80\textwidth]{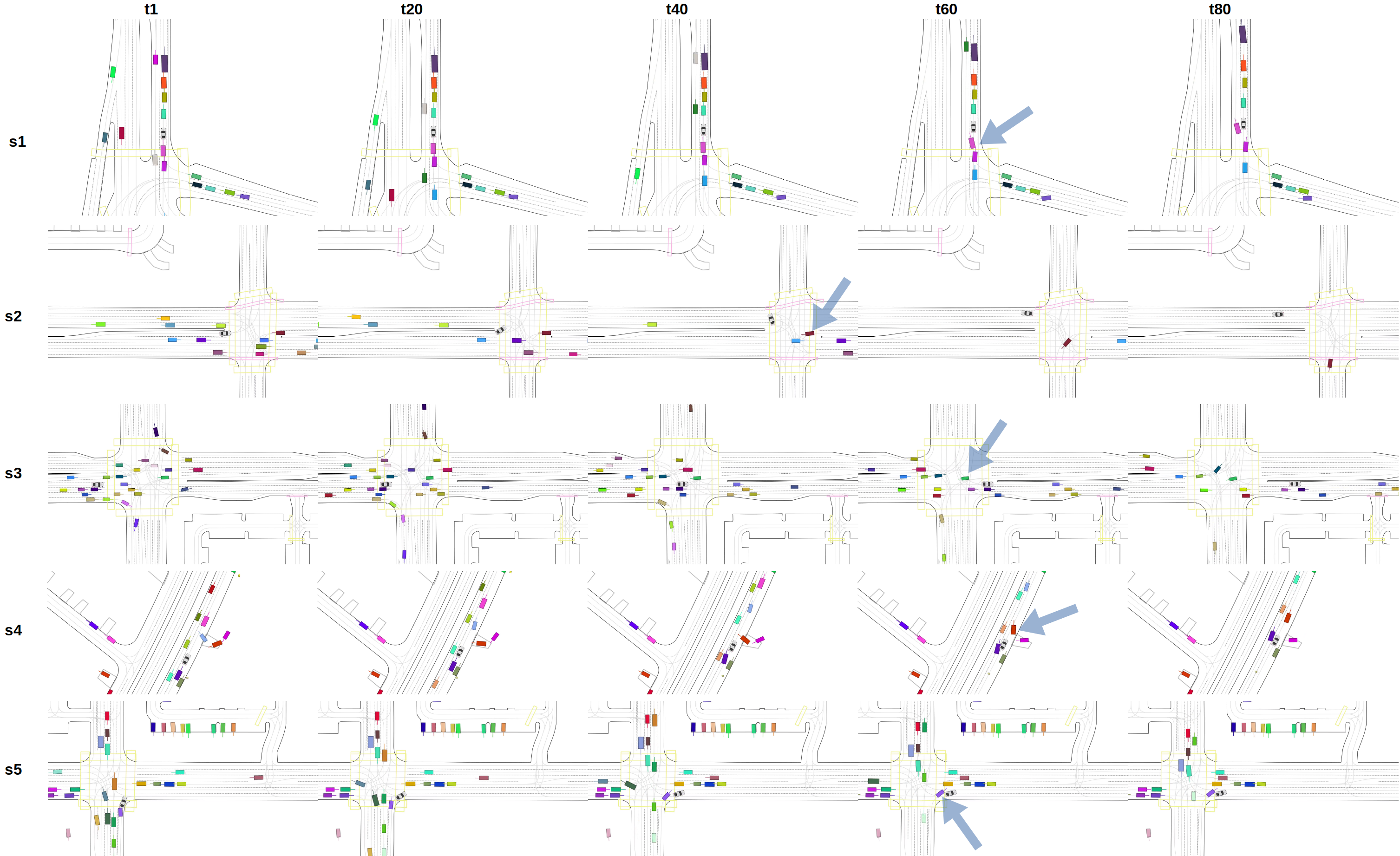}}
    \caption{Five simulated scenarios featuring reactive environment agents, each with five timesteps rendered. Scenario 1 depicts a congested right lane where an agent, indicated by the arrow, attempts to overtake the ADV. Scenario 2 and 3 showcase vehicles at intersections, waiting for the oncoming traffic to clear before making an unprotected left turn. Scenario 4 shows the ADV slows down to yield to a car merging onto the main road from a driveway. The car behind the ADV then changes to the left lane and overtakes the autonomous vehicle. Lastly, scenario 5 demonstrates the ADV executing a slow right-turn, resulting in the agent behind it having to slow down or stop.}
    \label{fig:qualitative_example2}
\end{figure*}

\begin{figure*}
    \centering
    \subfloat{\includegraphics[width=0.73\textwidth]{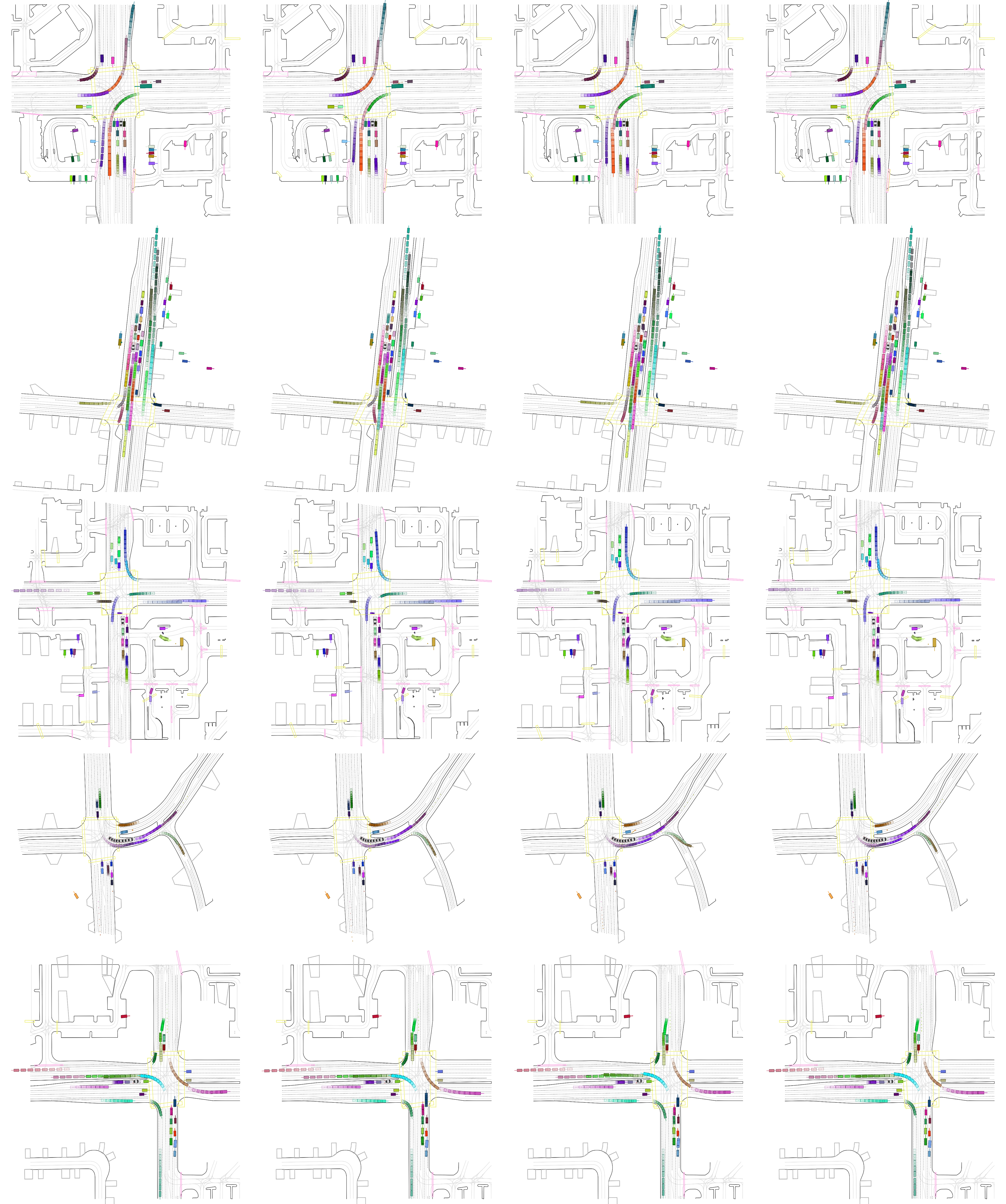}}\\
    \caption{Qualitative simulation results of several complex intersection scenarios. Each row shows 4 out of the 32 simulations for a scenario. }
    \label{fig:qualitative_results}
\end{figure*}

\noindent \textbf{Qualitative results}. In Figure~\ref{fig:qualitative_example1}, we present a scenario demonstrating the multi-modal behavior of an agent. Figure~\ref{fig:qualitative_example2} features five simulated scenarios showcasing reactive environment agents. These agents exhibit a wide variety of behaviors including yielding, overtaking, pausing for unprotected left turns, and engaging with the ADV. Qualitative simulation results of several intersection scenes with agents undertaking a wide variety of maneuvers are provided in Figure~\ref{fig:qualitative_results}. Due to the complexity of these scenes, it is impossible for heuristic-based models that encode traffic rules to simulate these realistic agents.

\section{Conclusion}
In this technical report, we have presented the Multiverse Transformer (MVTA) framework which produces parallel universes for the application of traffic agents simulation. It achieved state-of-the-art performance and ranks the 1\textsuperscript{st} place in the Waymo Open Sim Agents Challenge 2023. We hope our work inspires further research in the area of simulation agents. In our upcoming research, we intend to investigate scene-centric simulation approaches for improving the degree of realism of simulations, and also explore the possibility of using diffusion/denoising-based approaches. 

{\small
\bibliographystyle{ieee_fullname}
\bibliography{egbib.bib}
}

\end{document}